%% file: main.tex
\documentclass{article}
\usepackage{arxiv}
\usepackage[T1]{fontenc}
\usepackage{amsmath,amssymb}
\usepackage{graphicx}
\usepackage{booktabs}
\usepackage{xcolor}
\usepackage{subcaption}
\usepackage{array}
\usepackage[authoryear,round]{natbib}
\usepackage[hidelinks]{hyperref}
\newcommand{\makecell}[1]{\begin{tabular}[b]{@{}c@{}}#1\end{tabular}}

\newcommand{\x}{$\times$}
\newsavebox{\fitbox}
\newcommand{\fittable}[1]{\sbox{\fitbox}{#1}\ifdim\wd\fitbox>\linewidth\resizebox{\linewidth}{!}{\usebox{\fitbox}}\else\usebox{\fitbox}\fi}


\makeatletter
\g@addto@macro\@floatboxreset{\linespread{1}\selectfont}
\makeatother

\title{Neighbourhood shortcuts in building damage assessment from very-high-resolution satellite imagery: diagnosis and a building-evidence plug-in}
\author{JianChong Guo \quad YuTing Wan\thanks{Corresponding author: wanyuting@whu.edu.cn} \quad Kefan Li \quad AiLong Ma \quad YanFei Zhong \\
\normalfont\small State Key Laboratory of Information Engineering in Surveying, Mapping, and Remote Sensing, \\
\normalfont\small Wuhan University, 430079, China \\
\normalfont\small jianchongguo@whu.edu.cn \quad wanyuting@whu.edu.cn \quad likefan@whu.edu.cn \\
\normalfont\small maailong007@whu.edu.cn \quad zhongyanfei@whu.edu.cn}
\date{}
\renewcommand{\shorttitle}{Neighbourhood shortcuts in building damage assessment}
\hypersetup{pdftitle={Neighbourhood shortcuts in building damage assessment from very-high-resolution satellite imagery: diagnosis and a building-evidence plug-in},pdfauthor={JianChong Guo, YuTing Wan, Kefan Li, AiLong Ma, YanFei Zhong}}
\begin{document}
\maketitle
\input{sections/abstract}

\input{sections/introduction}
\input{sections/related}
\input{sections/data}
\input{sections/diagnosis}
\input{sections/method}
\input{sections/results}
\input{sections/discussion}

\section*{CRediT authorship contribution statement}
\textbf{JianChong Guo:} Conceptualization, Methodology, Software, Validation, Formal analysis, Investigation, Data curation, Visualization, Writing -- original draft, Writing -- review \& editing.
\textbf{YuTing Wan:} Supervision, Writing -- review \& editing.
\textbf{Kefan Li:} Visualization, Writing -- original draft.
\textbf{AiLong Ma:} Project administration.
\textbf{YanFei Zhong:} Funding acquisition.

\section*{Declaration of competing interest}
The authors declare that they have no known competing financial interests or personal relationships that could have appeared to influence the work reported in this paper.

\section*{Data availability}
xBD, xFBD and BRIGHT are publicly available from their original sources, as are the released weights of the benchmark models. The \.{I}slahiye images are from the Maxar Open Data Program under the CC BY-NC 4.0 licence, and the building instances and damage labels of the scene are from BRIGHT. The code of the evaluation protocol and of the plug-in is available at \url{https://github.com/SummerTide/neighbourhood-shortcut-bda}.

\section*{Acknowledgements}
This study was partially supported by the National Key Research and Development Program of China (Grant No. 2024YFC3015403), the National Natural Science Foundation of China (Grant Nos. 42671489, 42301418, and 42325105), the Hubei Provincial Natural Science Foundation of China (Grant No. 2026AFA013), and the LIESMARS Special Research Funding.

\section*{Declaration of generative AI and AI-assisted technologies in the manuscript preparation process}
During the preparation of this work, the authors used Codex (OpenAI) to support code development, data analysis, and language editing. After using this tool, the authors reviewed and edited the content as needed and take full responsibility for the content of the published article.

\bibliographystyle{elsarticle-harv}
\bibliography{refs}

\end{document}

%% file: sections/abstract.tex
\begin{abstract}
Building damage assessment (BDA) from very-high-resolution imagery supports disaster response, and deep models now achieve high overall accuracy on public benchmarks. Because damage is spatially clustered, most buildings share the damage level of their neighbours, so overall scores are dominated by such buildings, whereas damaged buildings among intact ones are rare and easily overlooked. A high overall score therefore cannot tell whether a model judges the building itself or relies on its neighbourhood as a shortcut, and this study tests which of the two current models do. We stratify buildings by their agreement with their neighbours, move buildings into new neighbourhoods with their pixels unchanged, and diagnose nine models covering the main paradigms of BDA. All models miss a third to half of the isolated damaged buildings, and in non-flood events they call 73--87\% of intact buildings damaged once these are moved into a damaged neighbourhood. We propose a building-evidence plug-in that adds two experts to an existing model without retraining it. One expert sees only the building, and the other sees its surroundings but is trained with seamlessly swapped neighbourhoods so that it learns not to trust them. With each of the nine models, the plug-in maintains or improves overall accuracy, reduces errors on buildings that disagree with their neighbours by 8--18\%, and cuts by 40--69\% the intact buildings that a damaged neighbourhood turns into damaged ones. It also raises the detection of synthetic isolated damage from 1--6\% to 13--39\% and carries over to optical--SAR data and to an unseen earthquake. High overall scores thus conceal a systematic reliance on the neighbourhood, which building-level evidence reduces at low cost.
\end{abstract}

%% file: sections/introduction.tex
\section{Introduction}
\label{sec:intro}

Building damage assessment (BDA) from very-high-resolution (VHR) satellite imagery is a routine input to disaster response, recovery planning and humanitarian reporting, and deep models now reach high overall scores on public benchmarks \citep{gupta2019xbd,zheng2021changeos}. Since the release of xBD \citep{gupta2019xbd}, the field has moved from pixel-wise segmentation ensembles \citep{durnov2020xview2} to object-level change detection \citep{zheng2021changeos}, graph networks over buildings \citep{hasan2026krgat}, change detectors pre-trained on synthetic change data \citep{zheng2024changen2} and state-space models \citep{chen2024changemamba}. Overall macro-F1 on xBD has converged at about 0.80, and the differences between recent methods are of the order of one point.

Damage is spatially autocorrelated, as Tobler's first law of geography leads one to expect \citep{tobler1970}, so that the damage of the neighbours is itself a good predictor of the damage of a building. Buildings close to each other share the hazard intensity, the construction type and the time of exposure, and in xBD the large majority of buildings have the same damage level as most of their neighbours. A model can therefore score well on the benchmark by judging the neighbourhood instead of the building. This would be a form of shortcut learning \citep{geirhos2020shortcut}, comparable to object classifiers that rely on the image background \citep{beery2018terra,xiao2021noise}, which work on the benchmark distribution but fail wherever the shortcut and the target disagree.

\begin{figure*}[!htbp]
\centering
\includegraphics[width=0.9\textwidth]{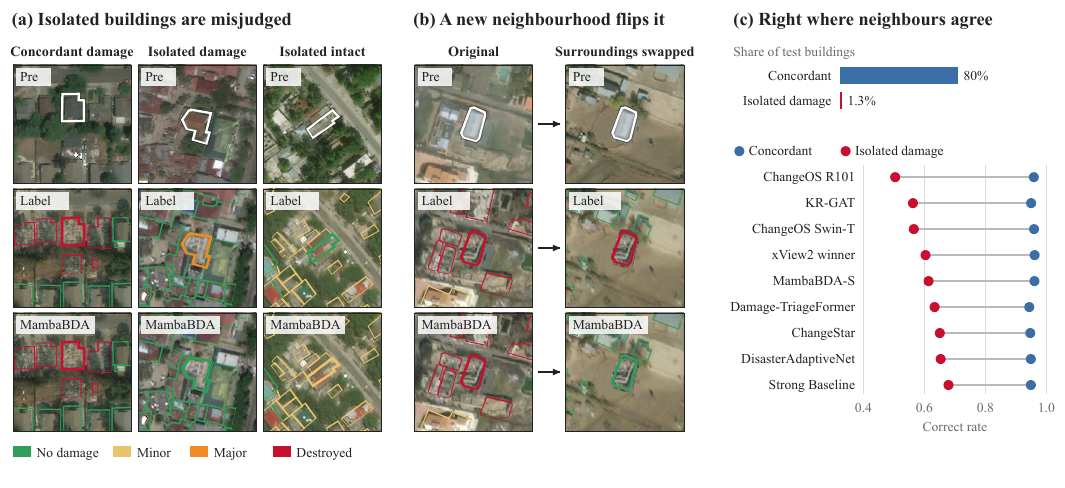}
\caption{Neighbourhood reliance of BDA models on the xBD test set. (a) Isolated buildings are misjudged. (b) Swapping the surroundings flips the decision. (c) Accuracy on concordant buildings and detection of isolated damaged buildings per model.}
\label{fig:problem}
\end{figure*}

Precisely because damage is usually clustered, its exceptions, damaged buildings among intact ones, are easily overlooked by models and benchmarks alike. Single structural failures after an earthquake, a house destroyed by a gas explosion, the edge of a tornado track and missile or artillery strikes in armed conflicts \citep{mueller2021monitoring,aimaiti2022war} all leave damaged buildings in intact streets. A model that follows the neighbourhood misses such a building, as Fig.~\ref{fig:problem} illustrates for a state-of-the-art model. It can do so and still rank first on a benchmark, because such buildings are rare there. In the non-flood events of xBD, only 10\% of the damaged buildings have mostly intact surroundings, against 43\% in the town of \.{I}slahiye after the 2023 Kahramanmara\c{s} earthquakes (Section~\ref{sec:conflict-data}).

The failure on isolated damage has been noticed before, mainly on synthetic examples and a small set of 45 real isolated buildings, and the proposed remedy was to retrain models with such examples. \citet{melamed2023xfbd} built xFBD by pasting a single damaged building into an intact scene, showed that models trained on xBD rarely detect it, and retrained a model on such composites. Retraining solves the synthetic test but does not make the model look at the building. In our experiments, the retrained model detects 88\% of the composites, yet its accuracy on real buildings that disagree with their neighbours does not improve (Section~\ref{sec:mitigations}). The model thus learns the appearance of the composites rather than to rely on the building, and each existing model would have to be retrained in the same way.

Meanwhile, recent methods use the neighbourhood more rather than less. Graph networks aggregate the features of neighbouring buildings on the premise that buildings rarely fail in isolation \citep{ismail2022towards,hasan2026krgat}, and state-space and transformer models extend the receptive field to the whole image \citep{chen2022dual,kaur2023large,chen2024changemamba}. These designs raise overall scores, but because the scores are dominated by buildings that agree with their neighbours, the benchmarks cannot tell whether a gain comes from reading the building better or from reading the street. A causal test of neighbourhood reliance on real buildings and across paradigms is therefore needed to know what current models have learned, together with a remedy that works with the models already in use instead of retraining each of them.

The contributions of this paper are threefold.
\begin{enumerate}
\item \textbf{An evaluation protocol for neighbourhood reliance.} Buildings are stratified by their agreement with the neighbourhood, and counterfactual tests move a building, with its pixels unchanged, onto open ground in a different neighbourhood.
\item \textbf{A diagnosis of nine models covering the main paradigms of BDA.} All of them rely on the neighbourhood, including those that take the building footprint as input. The failures have two separable causes, a neighbourhood shortcut and the rarity and distinct appearance of isolated damage, while in floods the surroundings belong to the label definition.
\item \textbf{A building-evidence plug-in.} Two light experts, one that sees only the building and one trained with seamless neighbourhood swaps, are combined with any existing model as a product of experts, according to the hazard type. On all nine models the plug-in keeps or improves overall macro-F1, raises the accuracy on discordant buildings and cuts the counterfactual flip rate by 40--69\%, and it carries over to optical--SAR data and to an unseen earthquake.
\end{enumerate}

Section~\ref{sec:related} reviews related work and Section~\ref{sec:data} the data and models. Section~\ref{sec:diagnosis} diagnoses the reliance of the models on the neighbourhood, Section~\ref{sec:method} derives the plug-in from the diagnosis, and Section~\ref{sec:results} tests it. Section~\ref{sec:discussion} concludes.

%% file: sections/related.tex
\section{Related work}
\label{sec:related}

\subsection{Building damage assessment}
Building damage has been mapped from remote sensing for decades, first by visual interpretation and change detection and more recently by deep learning \citep{dong2013comprehensive,guo2026disasterkd,wang2026dora}. SAR intensity and coherence changes reveal damage regardless of cloud cover and daylight \citep{matsuoka2004use,plank2014rapid,ge2020review}, optical and SAR images have been combined to assess single buildings \citep{brunner2010earthquake}, and oblique aerial images add three-dimensional cues of structural failure \citep{vetrivel2018disaster}. Benchmarks for BDA pair pre- and post-event images with per-building damage labels, first from optical VHR imagery and more recently across sensors. xBD \citep{gupta2019xbd} provides optical image pairs from 19 events with about 850,000 building polygons and four damage levels defined by the Joint Damage Scale, and remains the reference benchmark. BRIGHT \citep{chen2025bright} pairs pre-event optical with post-event SAR images, with three damage levels and instance-level building labels. DisasterM3 \citep{wang2025disasterm3} extends the setting to more events, sensors and tasks.

Methods differ in how they represent a building, but all of them see its surroundings. Pixel-wise segmentation networks, typically Siamese encoder--decoders applied to the image pair, predict a damage map that is then aggregated over each footprint. Most of them fuse pre- and post-event features in Siamese networks \citep{xu2019building,weber2020building,bai2020pyramid}, add attention across the two dates \citep{hao2021attention,wu2021building,shen2022bdanet} or segment and classify buildings jointly \citep{gupta2021rescuenet}. Transformer encoders model long-range dependencies across the whole tile \citep{chen2022dual,kaur2023large,wang2026earthvl,wang2026cityvlm}. The winning xView2 solution is an ensemble of such networks \citep{durnov2020xview2}, which was later simplified into a strong single-model baseline \citep{gerard2024strong} and extended with disaster-type conditioning \citep{hafner2025dan}. Object-level change detection localises buildings and classifies them jointly \citep{zheng2021changeos}. Crop-based classifiers judge each building from a patch around its footprint \citep{nex2019structural}, also with foundation-model backbones \citep{triageformer2026}, and graph networks pass messages between neighbouring buildings \citep{ismail2022towards,hasan2026krgat}. Recent work trains change detectors from single-temporal images \citep{zheng2021change}, pre-trains them on synthetic image pairs produced by a generative change model \citep{zheng2024changen2} or uses state-space models whose receptive field covers the whole image \citep{chen2024changemamba}. Models trained on xBD also lose accuracy on hazards and regions not seen in training, and their in-distribution scores say little about this loss \citep{nex2019structural,valentijn2020multi,benson2020assessing}. The surroundings enter through the receptive field, the crop or the graph, and none of these methods is evaluated on buildings that disagree with their surroundings. Damage in armed conflicts, where isolated damage is common, has been mapped from VHR optical images with machine learning \citep{mueller2021monitoring} and at national scale from Sentinel-1 intensity and coherence time series \citep{aimaiti2022war,dietrich2025open,scher2025nationwide}.

\subsection{Context and shortcut learning}
Deep networks exploit cues that are correlated with the label in the training distribution but not causally related to it, and counterfactual tests that change only the context have exposed this in natural images, but rarely in damage assessment. In object recognition, background and co-occurrence cues explain a large part of classifier decisions \citep{geirhos2020shortcut,beery2018terra,xiao2021noise}. Detectors fail when an object is transplanted into an unusual scene \citep{rosenfeld2018elephant}, benchmarks carry biases of their own \citep{torralba2011unbiased}, and the dependence on context can be measured and reduced by controlling what a model sees \citep{shetty2019not,singh2020dont}. Methods against spurious correlations judge a model by its worst group, in which the cue and the label disagree, and improve that group by reweighting it during training \citep{sagawa2020distributionally} or by retraining only the last layer \citep{kirichenko2023last}. In remote sensing and ecology, spatial autocorrelation is known to inflate accuracy when test samples lie close to training samples \citep{roberts2017cross,ploton2020spatial,meyer2021predicting,karasiak2022spatial,kattenborn2022spatially}. Spatial autocorrelation \citep{tobler1970} thus makes context both informative and a potential shortcut, because neighbouring pixels and objects usually share their class, which helps on average and hurts on the exceptions. For BDA the exceptions are isolated damaged buildings, which are easily overlooked because they are rare.

\subsection{Approaches to isolated damage}
Existing remedies change the training data with synthetic composites and are evaluated on composites of the same kind, so that it remains open whether they reduce reliance on the real neighbourhood. \citet{melamed2023xfbd} introduced xFBD, in which damaged buildings are blended into intact scenes, both as a test and as a training augmentation. Copy-paste augmentation \citep{dwibedi2017cut,ghiasi2021copypaste} and Poisson blending \citep{perez2003poisson} are generic tools to decorrelate objects from their surroundings. Products of experts have been used against known biases in natural language inference, where a bias-only model is combined with the main model during training so that the main model learns what the bias cannot explain \citep{clark2019dont,he2019unlearn,karimimahabadi2020end}. Our plug-in also multiplies experts, but at inference and with experts that carry the evidence of the building rather than the bias. Our work differs in three respects. The diagnosis is causal, on real buildings and across nine models. The remedy does not retrain the existing model and applies to any of them. Its effect is measured on real discordant buildings and in counterfactual tests as well as on xFBD. On these measures, the xFBD augmentation solves its own synthetic test but leaves real discordant buildings largely unchanged (Section~\ref{sec:mitigations}).

%% file: sections/data.tex
\section{Data and models}
\label{sec:data}

\subsection{xBD}

xBD is the main benchmark of this study, and its official test split contains 53,850 labelled buildings. xBD \citep{gupta2019xbd} contains pre- and post-event Maxar WorldView image pairs of 1024\x1024 pixels at about 0.3--0.8\,m from 19 events, including earthquakes, tsunamis, floods, wildfires, volcanic eruptions and wind storms. Building polygons are drawn on the pre-event image, and each building receives one of the four levels of the Joint Damage Scale (Table~\ref{tab:jds}). Train and tier3 are used for training, hold for model selection and test for all reported results.

\begin{table}[!htbp]
\centering
\caption{The Joint Damage Scale of xBD, quoted from \citet{gupta2019xbd}.}
\label{tab:jds}
\footnotesize
\begin{tabular}{lp{0.68\linewidth}}
\toprule
Level & Structure description \\
\midrule
0 No damage & Undisturbed. No sign of water, structural or shingle damage, or burn marks. \\
1 Minor damage & Building partially burnt, water surrounding structure, volcanic flow nearby, roof elements missing, or visible cracks. \\
2 Major damage & Partial wall or roof collapse, encroaching volcanic flow, or surrounded by water/mud. \\
3 Destroyed & Scorched, completely collapsed, partially/completely covered with water/mud, or otherwise no longer present. \\
\bottomrule
\end{tabular}
\end{table}

The Joint Damage Scale defines flood damage through the surroundings, so that in flood events the neighbourhood is part of the label. Water surrounding the structure is minor damage, a building surrounded by water or mud is major damage, and one partially or completely covered with water or mud is destroyed. Statements about neighbourhood reliance are therefore made on the non-flood events, and flood events are treated separately (Sections~\ref{sec:floods} and~\ref{sec:hazard}).

\subsection{xFBD}
xFBD is a synthetic test of isolated damage on which none of our models is trained. \citet{melamed2023xfbd} take the pre-event image of a tile of the 2018 Palu (Sulawesi) earthquake as the post-event scene and blend into it, with Poisson blending, a single damaged building from the real post-event image. The pasted building keeps its damage level and all other buildings are labelled intact. Its 1,466 test images are built from xBD test tiles, so that none of their buildings is seen in training. We use xFBD to test the detection of synthetic isolated damage and to compare our method with the published augmentation.

\subsection{BRIGHT}
BRIGHT extends the evaluation to multimodal data, with pre-event optical and post-event SAR images. BRIGHT \citep{chen2025bright} covers disaster events worldwide at sub-metre resolution and labels each building as intact, damaged or destroyed. We use the instance labels released for its 2026 challenge and the standard split. Scenes without instance labels or without a redistributable pre-event image are left out, and 10\% of the training scenes are held out for model selection. This gives 2,099 training, 233 validation and 697 test scenes, and the test scenes contain 58,438 buildings, of which 52,803 are intact, 2,070 damaged and 3,565 destroyed. The authors released five benchmark models trained on the standard split, which we use as existing models.

\subsection{The 2023 earthquake in \.{I}slahiye}
\label{sec:conflict-data}
\.{I}slahiye, a town in Gaziantep province, was heavily damaged by the Kahramanmara\c{s} earthquakes of 6 February 2023 and serves as an unseen test scene. We use a WorldView-2 image of 27 December 2022 and a WorldView-3 image of 7 February 2023, released as analysis-ready RGB visual products in the Turkey and Syria Earthquake 2023 collection of the Maxar Open Data Program \citep{maxar2023opendata} and resampled to 0.5\,m, together with the building instances of BRIGHT \citep{chen2025bright} that cover the town. BRIGHT labels each building as intact, damaged or destroyed, so we merge the minor and major damage of xBD into one damaged class. We score the 2,496 complete instances, of which 2,320 are intact, 104 damaged and 72 destroyed, and 75 of the 176 damaged or destroyed buildings stand among intact neighbours. The post-event image is offset from the footprints by up to 5\,m, and we shift it onto the pre-event image by whole pixels. None of the models or experts has seen this event during training. For the large-scene analysis we select two subregions that show the two directions of neighbourhood errors, the town centre, a 768\,m square with 620 buildings of which 49 are destroyed and 40 damaged, and the snow-covered outskirts, a 640\,m square with 38 intact buildings.

\subsection{Nine models covering the main paradigms of BDA}
\label{sec:models}

We compare nine models from eight published methods that cover the main paradigms of Section~\ref{sec:related} (Table~\ref{tab:models}). The first group localises buildings itself and receives the footprints only for scoring, whereas the second group also takes the footprints as input. All models are trained on train and tier3, with model selection on the hold split. On BRIGHT, the existing models are the five benchmark models released by its authors, UNet \citep{ronneberger2015unet}, DeepLabV3+ \citep{chen2018encoder}, SiamAttnUNet \citep{hao2021attention}, SiamCRNN \citep{chen2020change} and DamageFormer \citep{chen2022dual}.

\begin{table*}[!htbp]
\centering
\caption{Models evaluated.}
\label{tab:models}
\footnotesize
\fittable{\begin{tabular}{lll}
\toprule
Group & Paradigm & Model \\
\midrule
Footprints for scoring only & Object-level change detection & ChangeOS, ResNet-101 and Swin-T \citep{zheng2021changeos,liu2021swin} \\
 & Pixel segmentation ensemble & xView2 first place \citep{durnov2020xview2} \\
 & Synthetic-change pre-training & ChangeStar with Changen2 \citep{zheng2021change,zheng2024changen2} \\
 & Pixel segmentation & Strong Baseline \citep{gerard2024strong} \\
 & Disaster-conditioned segmentation & DisasterAdaptiveNet \citep{hafner2025dan} \\
 & State-space change detection & MambaBDA-Small \citep{chen2024changemamba} \\
\midrule
Footprints also as input & Footprint-conditioned crop & Damage-TriageFormer \citep{triageformer2026} \\
 & Building graph network & KR-GAT \citep{hasan2026krgat} \\
\midrule
BRIGHT & Multimodal benchmark models & UNet, DeepLabV3+, SiamAttnUNet, SiamCRNN, \\
 & & DamageFormer \citep{chen2025bright} \\
\bottomrule
\end{tabular}}
\end{table*}

%% file: sections/diagnosis.tex
\section{Diagnosing neighbourhood reliance}
\label{sec:diagnosis}
\label{sec:protocol}

We diagnose the nine models in three steps, each a test followed directly by what it shows. All models are first scored per building on the same footprints (Section~\ref{sec:task}). Grouping the buildings by their agreement with their neighbours then shows where the models fail (Section~\ref{sec:strata}), and moving buildings into new neighbourhoods shows why (Section~\ref{sec:cf}). Together the two tests point to two causes and one exception, floods, which Section~\ref{sec:mechanism} separates and which shape the design of the plug-in.

\subsection{Per-building task and scoring}
\label{sec:task}
We study per-building damage classification given building footprints, the setting of operational damage mapping and of footprint-based methods. Operational products report damage on pre-event building inventories \citep{unosat,haste2026}, for which global footprint layers are openly available \citep{msbuildings,sirko2021openbuildings,osm}, and recent methods classify buildings given their polygons \citep{triageformer2026,hasan2026krgat}. The setting separates damage classification from localisation, which our question, whether a decision follows the building or its neighbours, requires.

All models are scored on the same footprints, so that every comparison is between models that know the same buildings. For pixel-wise models we average the predicted class probabilities over each reference footprint and take the most probable damage level, which is how their per-building output is formed in practice. Crop classifiers and graph models output one probability vector per building. Models that also take the footprint as input form a separate group (Section~\ref{sec:models}). Our numbers are therefore not comparable with the xView2 leaderboard, which scores localisation and classification jointly at the pixel level, and like \citet{hasan2026krgat} we do not compare per-building scores with leaderboard scores.

\subsection{Neighbourhood strata reveal a shared failure on discordant buildings}
\label{sec:strata}
\label{sec:shared}

Buildings are stratified by their agreement with their neighbours, because only buildings that disagree with the neighbourhood can show whether a model looks at the building. The neighbours of a building are the other buildings of the same image whose centroids lie within 64 pixels, which is about 30\,m at the typical xBD resolution of 0.5\,m. The strata play the role of local indicators of spatial association for a categorical label, which flag observations that differ from their neighbours \citep{anselin1995lisa}. Table~\ref{tab:strata} defines the strata and gives their sizes. Isolated damaged, isolated intact and level-discordant buildings together form the \emph{discordant} buildings, which make up 3.6\% of the xBD test buildings.

\begin{table*}[!htbp]
\centering
\caption{Neighbourhood strata and their sizes on the test sets.}
\label{tab:strata}
\footnotesize
\begin{tabular}{l>{\raggedright\arraybackslash}p{0.5\linewidth}rr}
\toprule
Stratum & Definition & xBD & BRIGHT \\
\midrule
Concordant & At least 75\% of the neighbours have the same damage level & 43,056 & 47,890 \\
Isolated damaged & Damaged, and fewer than 25\% of the neighbours are damaged & 704 & 404 \\
Isolated intact & Intact, and at least 75\% of the neighbours are damaged & 544 & 229 \\
Level discordant & Damaged, at least 25\% of the neighbours are damaged, and fewer than 25\% share its level & 694 & 195 \\
Mixed & None of the above & 2,216 & 1,131 \\
Without neighbours & No building within 64 pixels & 6,636 & 8,589 \\
\midrule
All & & 53,850 & 58,438 \\
\bottomrule
\end{tabular}
\end{table*}

Each stratum asks a different question, and we report one measure for each in addition to macro-F1, the unweighted mean of the per-level F1 scores over all buildings, in which rare damage levels count as much as frequent ones. \emph{Concordant accuracy}, the exact-level accuracy on concordant buildings, shows whether a model is accurate when the building and its neighbours point to the same answer, which a shortcut also achieves. \emph{Isolated damage detected}, the share of isolated damaged buildings predicted as any damage level, shows whether a model detects damage that its neighbours do not suggest. \emph{Isolated intact kept}, the share of isolated intact buildings predicted intact, shows whether a model keeps an intact building intact among damaged neighbours. A model that follows the neighbourhood fails on both, in opposite directions, and a model can raise either one at the expense of the other by leaning towards damaged or intact. We therefore report both and their mean, the \emph{discordant balanced accuracy}, which rises only if both improve and which a model that calls every building damaged, or every building intact, cannot raise above 0.5. \emph{Level-discordant accuracy}, the exact-level accuracy on level-discordant buildings, shows whether a model copies the damage level of its neighbours. Mixed buildings and buildings without neighbours carry no clear neighbourhood signal and enter only macro-F1.

The nine models differ by little overall but fail alike on discordant buildings (Table~\ref{tab:baselines}). Their macro-F1 spans only 0.786--0.822 and their accuracy on concordant buildings 0.944--0.961. Every model misses a third to half of the isolated damaged buildings, with detection rates from 0.50 for ChangeOS R101 to 0.68 for Strong Baseline. Models that detect more isolated damage keep fewer isolated intact buildings intact, 0.57--0.62 for ChangeStar, Strong Baseline and DisasterAdaptiveNet against 0.71--0.78 for the others, so that the discordant balanced accuracy stays within 0.61--0.67 for all of them. Level-discordant buildings, damaged buildings among neighbours of a different damage level, receive the correct level in only 38--47\% of the cases. A higher macro-F1 does not mean better results on discordant buildings, and the winning solution of the xView2 challenge, which has the highest macro-F1 of the nine models at 0.822, ranks only fifth on discordant buildings at 0.659. Knowing which building to judge does not help either, since the two models that take the footprint as input reach a discordant balanced accuracy of 0.669--0.672, only slightly above the models that do not, at 0.612--0.663.

\begin{table*}[!htbp]
\centering
\caption{Diagnosis of the nine models on the xBD test set. The neighbourhood strata are defined in Table~\ref{tab:strata} and the counterfactual relocation in Section~\ref{sec:cf}, and the arrows give the better direction.}
\label{tab:baselines}
\footnotesize
\setlength{\tabcolsep}{3pt}
\fittable{\begin{tabular}{lcccccccc}
\toprule
 & Overall & \multicolumn{4}{c}{Neighbourhood strata} & \multicolumn{2}{c}{Counterfactual relocation} & Synthetic \\
\cmidrule(lr){2-2}\cmidrule(lr){3-6}\cmidrule(lr){7-8}\cmidrule(lr){9-9}
Model & \makecell{Macro-F1\\$\uparrow$} & \makecell{Concordant\\accuracy $\uparrow$} & \makecell{Isolated damage\\detected $\uparrow$} & \makecell{Isolated intact\\kept $\uparrow$} & \makecell{Discordant\\balanced\\accuracy $\uparrow$} & \makecell{Damaged kept\\among intact $\uparrow$} & \makecell{Intact called\\damaged among\\damaged $\downarrow$} & \makecell{xFBD\\detected $\uparrow$} \\
\midrule
\input{tables/tab_baselines.tex} 
\bottomrule
\end{tabular}}
\end{table*}

\subsection{Counterfactual relocation shows that decisions follow the neighbourhood}
\label{sec:cf}
\label{sec:cfdiag}
Stratified metrics show where models fail but not why, because an isolated damaged building may simply be harder to see than a concordant one. We therefore move buildings into a different neighbourhood, with their pixels unchanged, and observe whether the decision follows. Moving a concordant damaged building among intact buildings creates an artificial isolated damaged building, and moving a concordant intact building among damaged ones an artificial isolated intact building, so that the two tests mirror the two discordant strata. For each direction we draw 500 target buildings from the xBD test split. Each target, with its footprint dilated by 4 pixels, is pasted in both images onto a vacant site of another test tile, at least 2 pixels away from any building. The site is chosen so that its labelled buildings within 64 pixels, at least four of them, are all intact or all with major damage or destroyed. No existing building is covered, so that the new surroundings keep their own labels. A further variant removes the target by inpainting and keeps only its surroundings. The target and its rim are pasted unchanged, so that the model sees exactly the pixels of the original building, and a building counts as called damaged when its predicted probability of any damage level exceeds 0.5. Targets and sites are drawn with a fixed seed and are identical for all models. Because flood damage is defined through the surroundings (Section~\ref{sec:floods}), the results are reported for non-flood events only.

Moving a building into another neighbourhood, with its pixels unchanged, changes the decision of every model, so the failure is caused by the surroundings rather than by the building being harder to see. On non-flood events, a concordant damaged building moved among intact buildings is still called damaged in only 11--23\% of the cases, so that the artificial isolated damaged buildings are missed even more often than the real ones. Conversely, a concordant intact building moved among heavily damaged buildings is called damaged in 73--87\% of the cases (Table~\ref{tab:baselines} and Fig.~\ref{fig:problem}b). When the target is removed and only its damaged surroundings remain, 56--89\% of the empty locations are still assigned a damage level.

\subsection{Two causes and the exception of floods}
\label{sec:mechanism}
\label{sec:floods}

The failures have two causes that can be separated, the neighbourhood shortcut and the rarity and different appearance of isolated damage. The neighbourhood is informative, since a \emph{tight} crop that keeps only the building and a 4-pixel rim, zoomed to the input size, lowers macro-F1 from 0.785 to 0.744 (Table~\ref{tab:resnet}). The tight model is almost insensitive to the counterfactual move. A concordant damaged building among intact buildings is still called damaged in 90--91\% of the cases, and an intact building among damaged ones is called damaged in only 4--5\%. It nevertheless misses 38--46\% of the isolated destroyed buildings, six to seven times its miss rate on concordant destroyed ones. The first cause is thus the shortcut, which the counterfactuals expose. The second is that isolated damage is rare, with 4,742 isolated damaged buildings in the training split, and may look different from the extended debris of concordant destruction. A remedy must address both, and Section~\ref{sec:method} answers each cause with one expert.

\begin{table*}[!htbp]
\centering
\caption{Ablation of the experts on the xBD test set, with a context crop classifier as the existing model. Level is the level-discordant accuracy, and the plug-in is in bold.}
\label{tab:resnet}
\footnotesize
\setlength{\tabcolsep}{3pt}
\fittable{\begin{tabular}{lcccccc}
\toprule
Configuration & Macro-F1 & Conc. & Iso.\ dmg. & Iso.\ int. & Disc. & Level \\
\midrule
\input{tables/tab_resnet.tex} 
\bottomrule
\end{tabular}}
\end{table*}

In floods, judging a building by its surroundings follows the label definition rather than a shortcut. The Joint Damage Scale defines flood damage by the water around the structure (Table~\ref{tab:jds}), so that the surroundings are part of the label. Removing the surroundings accordingly costs much more on floods than elsewhere, and the tight crop loses 0.09 of macro-F1 on flood events against 0.01 on the other events. The plug-in therefore treats flood events differently (Section~\ref{sec:hazard}).

%% file: tables/tab_baselines.tex
\multicolumn{9}{l}{\emph{Footprints used for scoring only (pixel outputs averaged over each footprint)}} \\
ChangeOS (R101) & 0.807 & 0.959 & 0.504 & 0.748 & 0.626 & 0.16 & 0.82 & 0.03 \\
ChangeOS (Swin-T) & 0.814 & 0.959 & 0.565 & 0.754 & 0.660 & 0.17 & 0.76 & 0.01 \\
xView2 1st place & 0.822 & 0.961 & 0.604 & 0.715 & 0.659 & 0.20 & 0.87 & 0.02 \\
ChangeStar (Changen2) & 0.800 & 0.947 & 0.651 & 0.574 & 0.612 & 0.11 & 0.86 & 0.01 \\
Strong Baseline & 0.795 & 0.949 & 0.679 & 0.607 & 0.643 & 0.21 & 0.78 & 0.05 \\
DisasterAdaptiveNet & 0.795 & 0.949 & 0.653 & 0.616 & 0.635 & 0.23 & 0.73 & 0.02 \\
MambaBDA-S & 0.813 & 0.961 & 0.614 & 0.713 & 0.663 & 0.16 & 0.78 & 0.06 \\
\midrule
\multicolumn{9}{l}{\emph{Footprints also used as input}} \\
Damage-TriageFormer & 0.786 & 0.944 & 0.634 & 0.710 & 0.672 & -- & -- & -- \\
KR-GAT & 0.793 & 0.950 & 0.562 & 0.776 & 0.669 & -- & -- & -- \\

%% file: tables/tab_resnet.tex
Context crop & 0.785 & 0.947 & 0.584 & 0.754 & 0.669 & 0.391 \\
Tight crop & 0.744 & 0.927 & 0.566 & 0.835 & 0.700 & 0.379 \\
Swap-trained & 0.765 & 0.934 & 0.674 & 0.720 & 0.697 & 0.449 \\
\midrule
Context $\times$ context (control) & 0.795 & 0.951 & 0.582 & 0.757 & 0.670 & 0.388 \\
Context $\times$ context $\times$ context (control) & 0.799 & 0.953 & 0.591 & 0.757 & 0.674 & 0.393 \\
\midrule
Context $\times$ tight & 0.791 & 0.949 & 0.566 & 0.810 & 0.688 & 0.402 \\
Swap-trained $\times$ tight & 0.782 & 0.943 & 0.620 & 0.811 & 0.716 & 0.437 \\
\textbf{Context $\times$ tight $\times$ swap-trained (ours)} & \textbf{0.796} & \textbf{0.950} & \textbf{0.607} & \textbf{0.795} & \textbf{0.701} & \textbf{0.426} \\

%% file: sections/method.tex
\section{A building-evidence plug-in}
\label{sec:method}

\begin{figure*}[!htbp]
\centering
\includegraphics[width=0.82\textwidth]{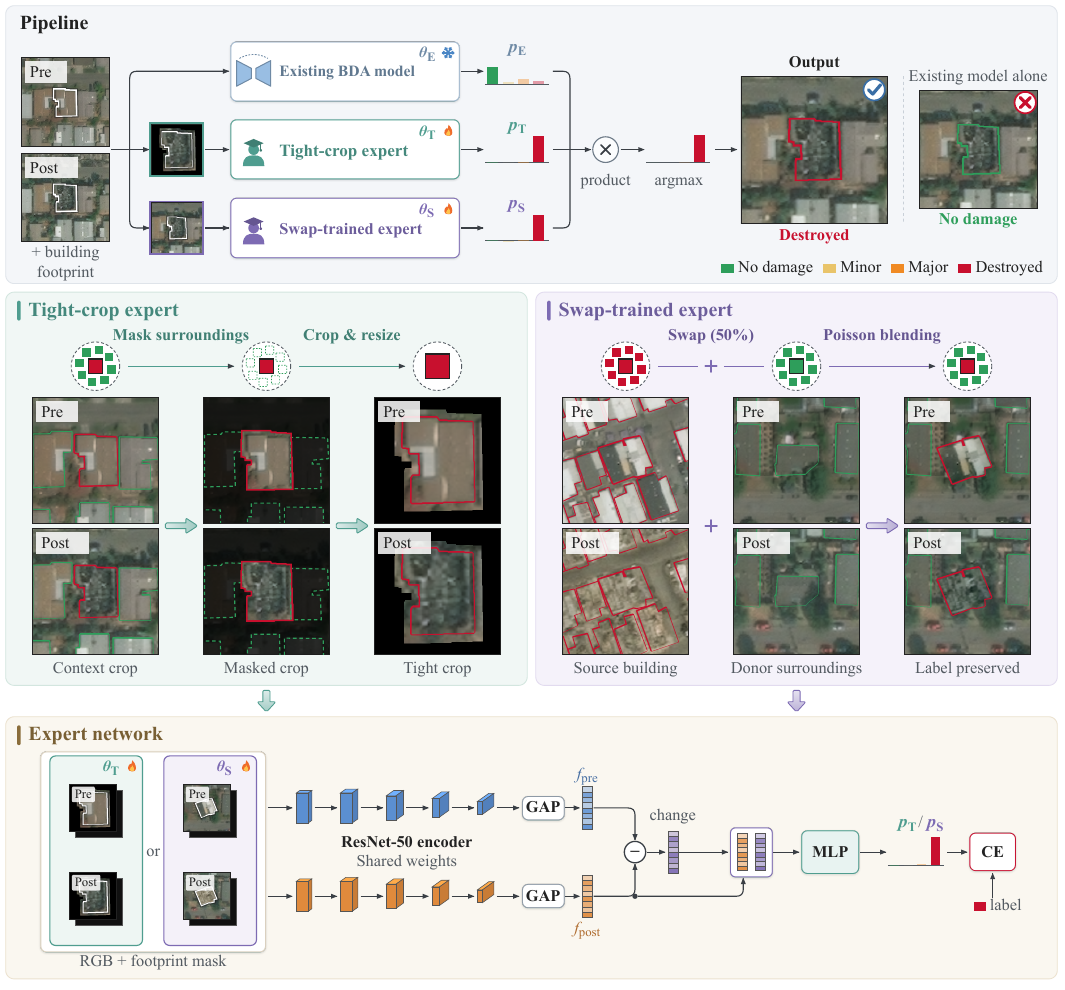}
\caption{Overview of the building-evidence plug-in. Top, the inference pipeline. Middle, the inputs of the tight-crop and the swap-trained expert. Bottom, the expert network shared by both.}
\label{fig:method}
\end{figure*}

The plug-in answers each finding of the diagnosis with one design choice and leaves the existing model unchanged (Fig.~\ref{fig:method}). Against the neighbourhood shortcut, a tight-crop expert judges the building without its surroundings (Section~\ref{sec:tight}). Against the rarity of isolated damage, a swap-trained expert sees the surroundings but is trained on swapped neighbourhoods, which also breaks the shortcut, and with discordant buildings oversampled (Section~\ref{sec:swap}). Since the neighbourhood carries real information, the two experts are combined with the existing model as a product of experts, which keeps the neighbourhood as evidence without letting it decide alone (Section~\ref{sec:poe}), and for floods, where the neighbourhood is part of the label, the existing model is used alone (Section~\ref{sec:hazard}).

Both experts are crop classifiers of the same architecture, a ResNet-50 that reads the pre- and post-event crops with the footprint as a fourth channel, and differ only in what they see and how they are trained. Each expert passes the pre- and post-event crops of 224\x224 pixels through a shared ResNet-50, classifies the concatenation of the post-event features and the feature difference with a two-layer MLP, and is trained with the cross-entropy loss and AdamW \citep{loshchilov2019decoupled}, at a learning rate and weight decay of $10^{-4}$ with a one-cycle schedule \citep{smith2019super}, a batch size of 128, 8 epochs and random rotations and flips. Each expert trains in under an hour on one NVIDIA A100 GPU with 40 GB of memory.

\subsection{Tight-crop expert for evidence from the building alone}
\label{sec:tight}
The tight-crop expert cannot use the neighbourhood, because it sees the building and almost nothing else. Its input is the footprint bounding box enlarged by 8 pixels and resized to 224\x224 pixels, with every pixel farther than 4 pixels from the footprint set to zero in both the pre- and post-event images. The zoom gives small buildings more pixels, and the 4-pixel rim keeps debris and water immediately around the building.

\subsection{Swap-trained expert for context without reliance on the neighbourhood}
\label{sec:swap}
The swap-trained expert sees the neighbourhood but is trained not to trust it, and with more weight on the rare discordant buildings. Its input is the same crop as that of an ordinary context crop classifier, the footprint bounding box enlarged by 32 pixels. During training, with probability 0.5, everything farther than $m$ pixels from the footprint is replaced in both images by the surroundings of another building of the same event, while the label stays that of the building. Half of the donor buildings are damaged, so that intact buildings also appear among damaged neighbours and damaged buildings among intact ones. The rim width $m$ is drawn uniformly between 2 and 12 pixels, and the building is blended into its new surroundings with Poisson image editing \citep{perez2003poisson}, so that no seam marks the swap. Swapping within the same event keeps the radiometry similar. The swap breaks the link between a building's label and its surroundings, which removes the shortcut. In addition, buildings of the discordant strata are sampled five times more often, which addresses their rarity.

\subsection{Product of experts with the existing model}
\label{sec:poe}
The three models are combined as a product of experts, so that a confident objection from one expert can overrule the others while agreement leaves the decision unchanged. Given the class probabilities $p_k(c \mid x)$ of the experts $k = 1, \dots, K$ for a building $x$ over the damage levels $c$, the product of experts \citep{hinton2002poe} predicts
\begin{equation}
\hat{c} = \arg\max_c \sum_{k=1}^{K} \log p_k(c \mid x).
\end{equation}
The plug-in uses $K = 3$ experts, namely the existing model, which can be any of the models of Section~\ref{sec:models}, the tight-crop expert and the swap-trained expert. A class wins only if no expert finds it very unlikely, so that a confident building-level ``damaged'' from the tight-crop expert can overrule a neighbourhood-driven ``intact'', and vice versa. The combination has no parameters to fit and does not modify the existing model, which is used as released. The two experts are trained once and serve every existing model.

\subsection{Hazard-dependent use of the plug-in}
\label{sec:hazard}
The plug-in is applied according to the hazard type, which is known in any damage mapping activation, and for floods the existing model is used alone. Flood damage is defined through the surrounding water (Section~\ref{sec:floods}), and experts built to discount the surroundings work against this definition. Without the hazard rule, the plug-in lowers the flood macro-F1 of five of the nine models, by up to 0.04. With it, flood buildings keep the existing model's prediction and all other hazards receive the plug-in, and the same rule applies to every dataset and modality.

%% file: sections/results.tex
\section{Experiments}
\label{sec:results}

The experiments test the plug-in against the diagnosis, one claim per section. An ablation with a context crop classifier as the existing model shows that the gain comes from the design of the two experts and not from combining three models (Section~\ref{sec:ablation}). Attached to the nine diagnosed models, the plug-in improves the strata on which Section~\ref{sec:strata} found them to fail (Section~\ref{sec:allmodels}), with any expert backbone and over a wide range of parameters (Section~\ref{sec:sensitivity}), and the counterfactual tests of Section~\ref{sec:cf} and Grad-CAM show that it does so by relying less on the neighbourhood (Section~\ref{sec:cfplugin}). A comparison with retraining on synthetic composites and the transfer to optical--SAR data test how general the gain is (Section~\ref{sec:bright}), and an unseen earthquake tests it on a whole town (Section~\ref{sec:conflict}).

\subsection{Contribution of each expert in an ablation}
\label{sec:ablation}

The ablation shows what each expert contributes, with a context crop classifier of the experts' architecture as the existing model (Table~\ref{tab:resnet}). Any combination of three models tends to improve on one model \citep{lakshminarayanan2017simple}, so we also report an ensemble control. The control replaces the two experts by two ordinary context crop classifiers with the same architecture, training and combination, but with neither the tight crop nor the swap. Whatever the plug-in gains beyond this control is due to the design of the experts.

Each expert alone trades overall accuracy for discordant buildings. The tight-crop expert keeps the most isolated intact buildings intact, 0.835, and the swap-trained expert detects the most isolated damaged buildings, 0.674, but they lose 4 and 2 points of macro-F1 against the context crop. Combined with the existing model, they recover the overall accuracy and keep the gain on discordant buildings, with a macro-F1 of 0.796 and a discordant balanced accuracy of 0.701. The ensemble control raises macro-F1 by the same amount, to 0.795--0.799, but leaves the discordant balanced accuracy where it was, at 0.670--0.674. The gain on discordant buildings therefore comes from the experts, not from combining three models. Without the existing model, the two experts reach the highest discordant balanced accuracy, 0.716, but lose 1.4 points of macro-F1 against the full plug-in, which shows why the existing model is kept.

\subsection{Gains on all nine diagnosed models}
\label{sec:allmodels}

Attached to each of the nine diagnosed models, the plug-in keeps or improves overall accuracy and improves every model on discordant buildings (Table~\ref{tab:plugin}). Macro-F1 rises by 0.003--0.021, and the accuracy on concordant buildings stays at its level. The discordant balanced accuracy rises by 2.6--6.6 points, which removes 8--18\% of the errors on discordant buildings. Most of the gain comes from isolated intact buildings that are no longer called damaged, whose share kept intact rises by 2.1--11.5 points. The level-discordant accuracy improves for eight of the nine models.

\begin{table*}[!htbp]
\centering
\caption{The plug-in on the nine models, on the xBD test set and on xFBD. Det.\ is the detection of the pasted damaged buildings and Int.\ the intact buildings around them kept intact. Bold marks the better value of each pair.}
\label{tab:plugin}
\footnotesize
\setlength{\tabcolsep}{3pt}
\fittable{\begin{tabular}{llcccccccc}
\toprule
 & & \multicolumn{6}{c}{xBD} & \multicolumn{2}{c}{xFBD} \\
\cmidrule(lr){3-8}\cmidrule(lr){9-10}
Model & & Macro-F1 & Conc. & Iso.\ dmg. & Iso.\ int. & Disc. & Level & Det. & Int. \\
\midrule
\input{tables/tab_plugin.tex} 
\bottomrule
\end{tabular}}
\end{table*}

On xFBD, which none of the models was trained on, the plug-in finds many of the pasted damaged buildings that the existing models miss. Alone, the seven models with xFBD predictions detect 1--6\% of them, and with the plug-in 13--39\%, while almost all intact buildings around them stay intact. The gain is largest for ChangeOS R101, MambaBDA-S and the xView2 winning solution, which reach 35--39\%, and smallest for ChangeOS Swin-T, ChangeStar and DisasterAdaptiveNet, which reach 13--16\%.

\begin{figure*}[!htbp]
\centering
\includegraphics[width=\textwidth]{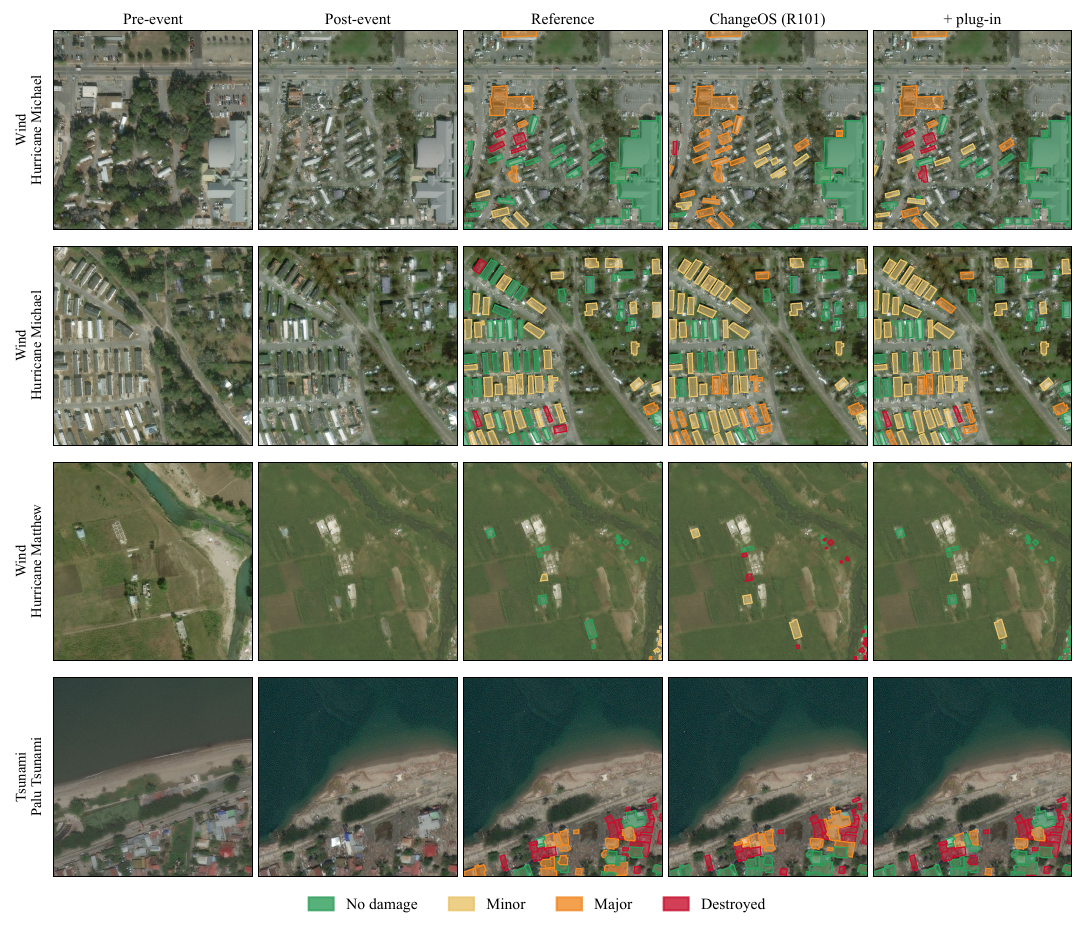}
\caption{ChangeOS (R101) alone and with the plug-in on test scenes of xBD.}
\label{fig:qualxbd}
\end{figure*}

The damage maps show how the plug-in corrects the existing model (Fig.~\ref{fig:qualxbd}). After Hurricane Michael, ChangeOS (R101) spreads the level of the neighbourhood over whole blocks of mobile homes, whereas the plug-in recovers the destroyed and the undamaged homes among them. After Hurricane Matthew, it removes destroyed predictions from intact buildings in open land, and after the Palu tsunami it separates destroyed houses from the damaged ones beside them.

\subsection{Robustness to the expert backbone and parameters}
\label{sec:sensitivity}

The gain does not depend on the backbone of the experts, and a stronger backbone increases it (Table~\ref{tab:backbone}). With both experts fine-tuned from the satellite DINOv3 ViT-L/16 \citep{simeoni2025dinov3} instead of the ResNet-50, with the same crops, swaps and training otherwise, the plug-in raises the discordant balanced accuracy of all nine models by 3.1--7.8 points, against 2.6--6.6 points with the ResNet-50 experts, and reaches the highest value for every model. Averaged over the nine models, the discordant balanced accuracy rises from 0.649 alone to 0.690 with the ResNet-50 experts and to 0.702 with the DINOv3 experts, while macro-F1 rises from 0.803 to 0.815 with either backbone. The plug-in thus benefits from progress in foundation models without retraining the existing model.

\begin{table}[!htbp]
\centering
\caption{The plug-in with experts on a ResNet-50 and on the satellite DINOv3 ViT-L/16 backbone, on the xBD test set. Alone gives the existing model, and + ResNet-50 and + DINOv3 the model with the plug-in on either backbone. Bold marks the best value of each measure for each model.}
\label{tab:backbone}
\footnotesize
\setlength{\tabcolsep}{3pt}
\fittable{\begin{tabular}{lcccccc}
\toprule
 & \multicolumn{3}{c}{Macro-F1} & \multicolumn{3}{c}{Discordant balanced accuracy} \\
\cmidrule(lr){2-4}\cmidrule(lr){5-7}
Model & Alone & + ResNet-50 & + DINOv3 & Alone & + ResNet-50 & + DINOv3 \\
\midrule
\input{tables/tab_backbone.tex} 
\bottomrule
\end{tabular}}
\end{table}

\begin{figure*}[!htbp]
\centering
\includegraphics[width=\textwidth]{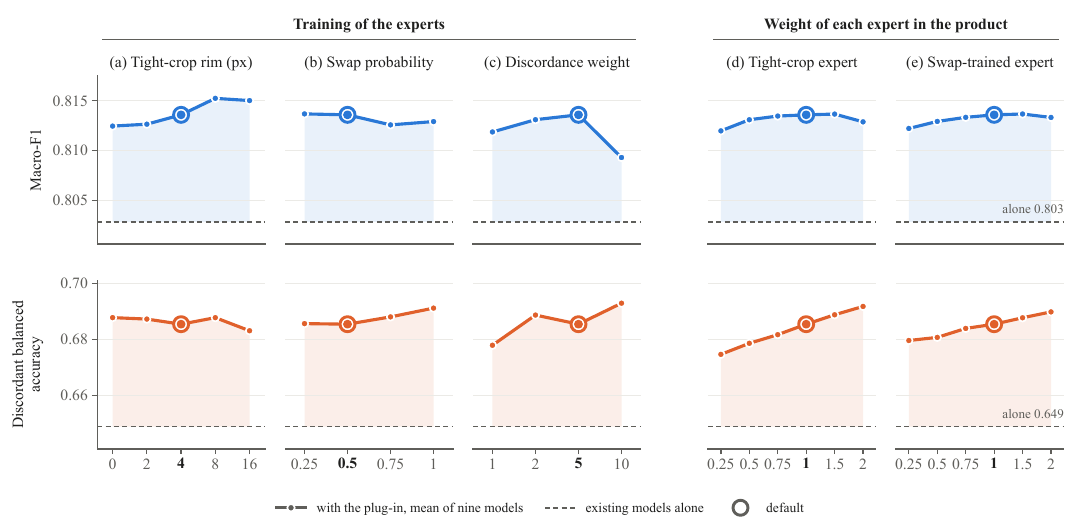}
\caption{Sensitivity of the plug-in to the parameters of the experts, as the mean over the nine models of macro-F1 and of the discordant balanced accuracy. Each panel changes one parameter and keeps the others at their default, marked by the ringed dot and the bold label. Panels (a) to (c) retrain an expert, and panels (d) and (e) weight the log-probability of an expert in the product. The dashed lines give the models alone, and the shading the gain of the plug-in over them.}
\label{fig:sensitivity}
\end{figure*}

The plug-in is also insensitive to its parameters (Fig.~\ref{fig:sensitivity}). Across a rim of 0 to 16 pixels for the tight-crop expert, a swap probability of 0.25 to 1 and a discordance weight of 1 to 10 for the swap-trained expert, the mean macro-F1 stays between 0.809 and 0.815 and the discordant balanced accuracy between 0.678 and 0.693, against 0.803 and 0.649 for the models alone. Only the largest discordance weight costs 0.5 points of macro-F1, because the rare discordant buildings then dominate the training. Scaling the log-probability of either expert by a weight from 0.25 to 2 changes the mean macro-F1 by at most 0.002, and the discordant balanced accuracy stays between 0.675 and 0.692, above the 0.649 of the models alone for every weight. A larger weight on either expert moves the decision further towards building evidence and raises the discordant balanced accuracy slightly, so the equal weights of the product of experts are a safe default rather than a tuned optimum.

\subsection{Lower reliance on the neighbourhood}
\label{sec:attention}
\label{sec:cfplugin}

The counterfactual tests of Section~\ref{sec:cf} show that the gain comes from a lower reliance on the neighbourhood (Table~\ref{tab:cfplugin}). An intact building moved among heavily damaged buildings is called damaged by the existing models in 73--87\% of the cases, and with the plug-in in only 27--44\%. A damaged building moved among intact buildings is still called damaged in 11--23\% of the cases by the existing models, and in 42--54\% with the plug-in. The plug-in thus cuts the intact buildings that a damaged neighbourhood turns into damaged ones by 40--69\% and keeps two to four times as many damaged buildings damaged among intact neighbours, while keeping the overall accuracy of the existing models.

\begin{table}[!htbp]
\centering
\caption{Counterfactual tests of the plug-in on the non-flood events of the xBD test set.}
\label{tab:cfplugin}
\footnotesize
\begin{tabular}{lcccc}
\toprule
 & \multicolumn{2}{c}{Kept $\uparrow$} & \multicolumn{2}{c}{Flip $\downarrow$} \\
\cmidrule(lr){2-3}\cmidrule(lr){4-5}
Model & Alone & + plug-in & Alone & + plug-in \\
\midrule
\input{tables/tab_cf_plugin.tex} 
\bottomrule
\end{tabular}
\end{table}

\begin{figure*}[!htbp]
\centering
\includegraphics[width=0.85\textwidth]{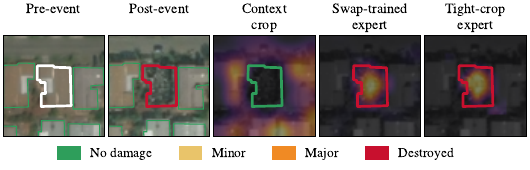}
\caption{Grad-CAM of the predicted class for an isolated destroyed building of the xBD test set. The outline gives the label on the post-event image and the prediction on each heat map.}
\label{fig:attention}
\end{figure*}

Inside the model, Grad-CAM \citep{selvaraju2017gradcam} of the predicted class shows the same shift, since the experts take their decision from the building (Fig.~\ref{fig:attention}). The context crop classifier, which has the experts' architecture but neither the tight crop nor the swap, calls an isolated destroyed building intact and draws this decision from the intact neighbours, with 1\% of its Grad-CAM on the building. The swap-trained expert sees the same crop but puts 70\% of its Grad-CAM on the building and calls it destroyed, as does the tight-crop expert with 83\%. The swap thus moves the evidence for the decision from the neighbours back to the building.

\subsection{Comparison with synthetic retraining and transfer to optical--SAR data}
\label{sec:mitigations}
\label{sec:bright}

The published remedy for isolated damage improves only its own synthetic test, whereas the plug-in improves real discordant buildings without retraining the existing model (Table~\ref{tab:mitigation}). \citet{melamed2023xfbd} retrain the existing model with synthetic single-damage composites, while the plug-in leaves it unchanged and adds the two experts. We compare both on a context crop classifier of the experts' architecture. The retrained model detects 88\% of the pasted buildings of xFBD, which is expected since it has seen composites of this kind during training. On real xBD buildings, however, it does not improve. It detects 2.3 points more isolated damage but keeps 2.8 points fewer isolated intact buildings intact, so that its discordant balanced accuracy stays at 0.666, and it loses 1.2 points of macro-F1. The plug-in detects the same additional isolated damage and keeps 4.1 points more isolated intact buildings intact, which raises the discordant balanced accuracy to 0.701, and it raises macro-F1 to 0.796. Retraining on composites thus teaches a model to find the composites, whereas evidence from the building itself carries over to real discordant buildings.

\begin{table}[!htbp]
\centering
\caption{Comparison with retraining on xFBD composites, on the xBD test set. Bold marks the best value of each column.}
\label{tab:mitigation}
\footnotesize
\fittable{\begin{tabular}{lcccc}
\toprule
 & Overall & \multicolumn{3}{c}{Discordant buildings} \\
\cmidrule(lr){2-2}\cmidrule(lr){3-5}
 & Macro-F1 & Iso.\ dmg. & Iso.\ int. & Disc. \\
\midrule
\input{tables/tab_mitigation.tex} 
\bottomrule
\end{tabular}}

\end{table}

The gain also carries over to another modality. On BRIGHT, where the post-event image is SAR, the plug-in improves all five benchmark models on discordant buildings and keeps or improves their overall accuracy (Table~\ref{tab:bright}). The discordant balanced accuracy rises by 1.5--3.4 points, and macro-F1 rises for four of the five models, by up to 1.0 point, and stays at its level for DamageFormer. Most of the gain comes from isolated intact buildings, 3.1--8.3 points more of which are kept intact, whereas the detection of isolated damage stays at 7--11\%. A single damaged building leaves a weak signal in SAR, and even the tight-crop expert alone detects only 12\% of them. On optical--SAR data, the plug-in thus removes a large part of the false alarms that a damaged neighbourhood causes, while detecting isolated damage in SAR remains open.

\begin{table*}[!htbp]
\centering
\caption{The plug-in on the five benchmark models of BRIGHT. Bold marks the better value of each pair.}
\label{tab:bright}
\footnotesize
\setlength{\tabcolsep}{3pt}
\fittable{\begin{tabular}{llcccccc}
\toprule
Model & & Macro-F1 & Conc. & Iso.\ dmg. & Iso.\ int. & Disc. & Level \\
\midrule
\input{tables/tab_bright.tex} 
\bottomrule
\end{tabular}}
\end{table*}

\begin{figure*}[!htbp]
\centering
\includegraphics[width=\textwidth]{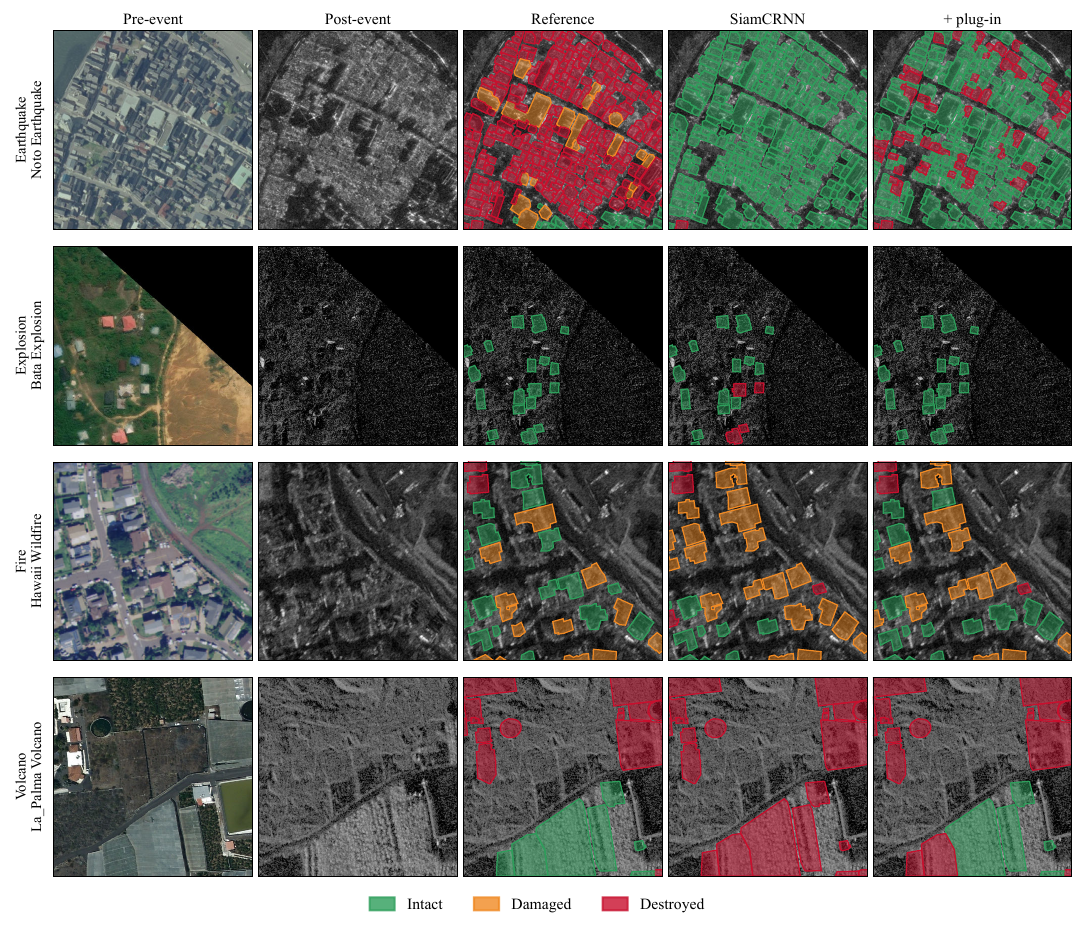}
\caption{SiamCRNN alone and with the plug-in on test scenes of BRIGHT. The post-event image is SAR.}
\label{fig:qualbright}
\end{figure*}

The damage maps of SiamCRNN show the same pattern (Fig.~\ref{fig:qualbright}). After the Bata explosion, the Hawaii wildfire and the La Palma eruption, the plug-in returns to intact most of the intact buildings that the model calls damaged or destroyed. After the Noto earthquake, where the model calls a destroyed town centre intact, the plug-in recovers part of the destroyed buildings.

\subsection{Large-scene application to an unseen earthquake}
\label{sec:conflict}

\begin{figure*}[!htbp]
\centering
\includegraphics[width=\textwidth]{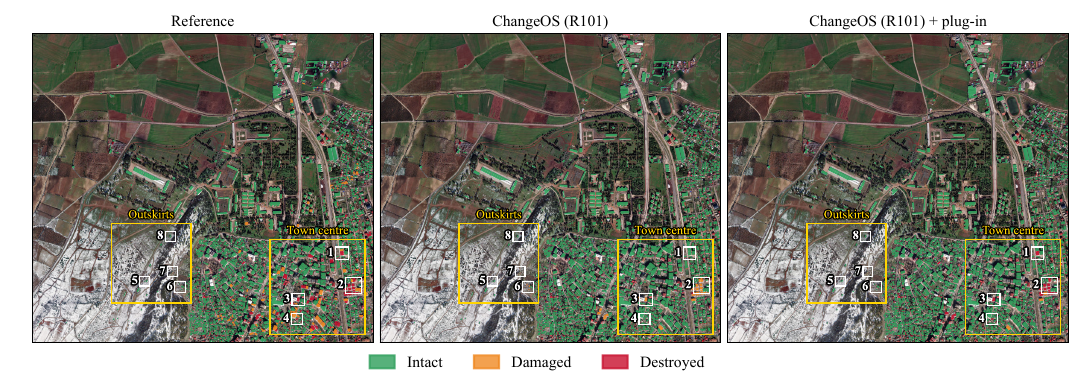}
\caption{ChangeOS (R101) alone and with the plug-in over the whole labelled scene of \.{I}slahiye, T\"{u}rkiye, after the 2023 Kahramanmara\c{s} earthquakes, about 2.8 by 2.5\,km with 2,496 buildings. The yellow boxes mark the town centre and the outskirts, and the numbered boxes the areas enlarged in Figs.~\ref{fig:turkey} and~\ref{fig:turkeysnow}.}
\label{fig:turkeyscene}
\end{figure*}

Applied to a whole town, the plug-in improves all three models tested, ChangeOS R101, MambaBDA-S and the xView2 winning solution (Fig.~\ref{fig:turkeyscene} and Table~\ref{tab:turkey}). We run them over the \.{I}slahiye scene, with the experts trained on xBD only, and score all 2,496 labelled buildings as well as the town centre and the snow-covered outskirts described in Section~\ref{sec:conflict-data}. Over the whole scene, the plug-in raises macro-F1 by 3.6--14.5 points and the recall of destroyed buildings by 9--31 points. At this scale the damage maps of the existing models are almost uniformly intact, whereas the plug-in brings out the collapsed buildings scattered through the town.

\begin{table}[!htbp]
\centering
\caption{The plug-in in \.{I}slahiye, over the whole labelled scene and its two sub-regions of Fig.~\ref{fig:turkeyscene}. Level gives the accuracy on level-discordant buildings, Destroyed the recall of destroyed buildings, and False alarms the number of the 38 intact buildings of the outskirts predicted as damaged. Bold marks the better value of each pair.}
\label{tab:turkey}
\footnotesize
\setlength{\tabcolsep}{3pt}
\fittable{\begin{tabular}{llccccccc}
\toprule
 & & \multicolumn{3}{c}{Whole scene} & \multicolumn{3}{c}{Town centre} & Outskirts \\
\cmidrule(lr){3-5}\cmidrule(lr){6-8}\cmidrule(lr){9-9}
Model & & Macro-F1 & Level & Destroyed & Macro-F1 & Level & Destroyed & False alarms \\
\midrule
\input{tables/tab_turkey.tex} 
\bottomrule
\end{tabular}}
\end{table}

\begin{figure*}[!htbp]
\centering
\includegraphics[width=\textwidth]{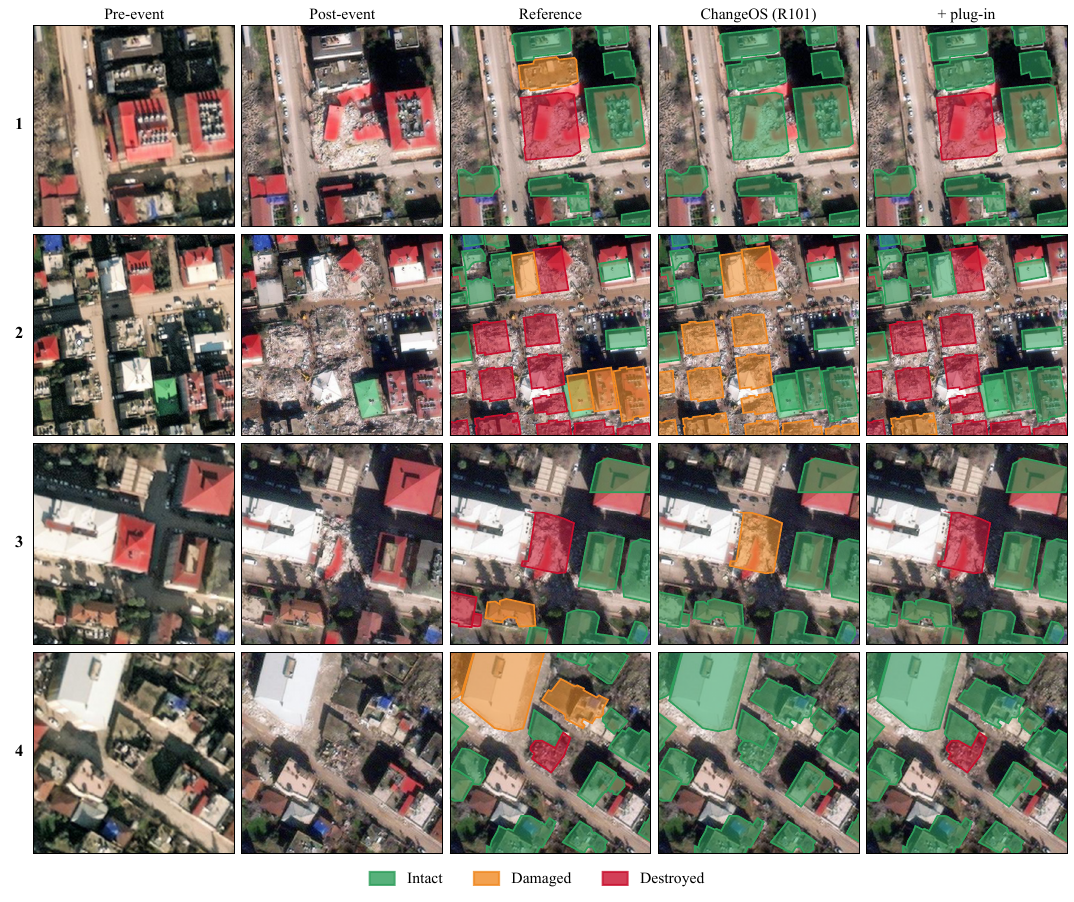}
\caption{ChangeOS (R101) alone and with the plug-in in the town centre of \.{I}slahiye, on the areas 1--4 marked in Fig.~\ref{fig:turkeyscene}.}
\label{fig:turkey}
\end{figure*}

In the town centre, the plug-in recovers collapsed buildings that the existing models assign to their neighbourhood (Fig.~\ref{fig:turkey}). Collapsed buildings stand next to damaged and intact ones, and the existing models call most of them intact or damaged, as they call their neighbours. The plug-in raises the recall of destroyed buildings by 10--33 points and the accuracy on buildings whose level differs from that of their neighbours from 0--17\% to 22--26\%, and it raises macro-F1 for all three models.

\begin{figure*}[!htbp]
\centering
\includegraphics[width=\textwidth]{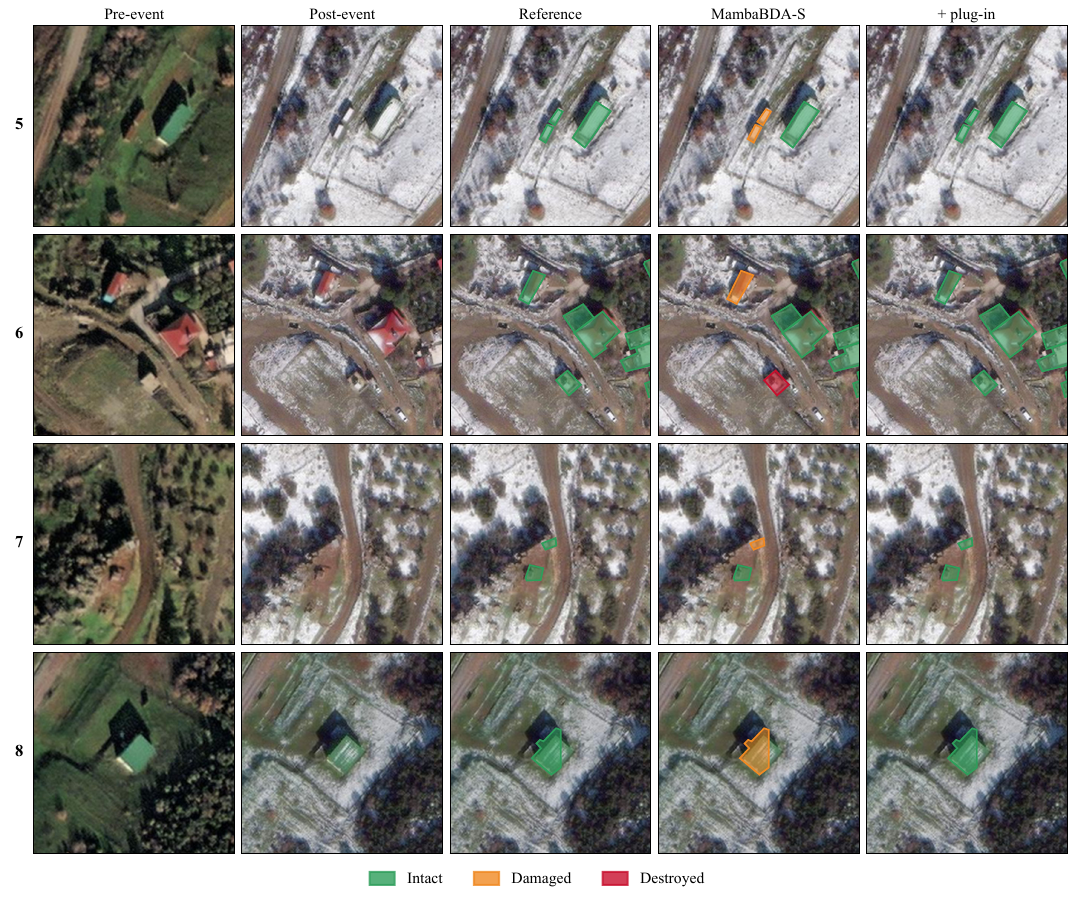}
\caption{MambaBDA-S alone and with the plug-in on the snow-covered outskirts of \.{I}slahiye, on the areas 5--8 marked in Fig.~\ref{fig:turkeyscene}.}
\label{fig:turkeysnow}
\end{figure*}

On the outskirts, the plug-in removes false alarms instead (Fig.~\ref{fig:turkeysnow}). Snow brightens the surroundings of intact buildings, and the existing models, which read the change around a building, call them damaged. The experts judge the building itself and give these buildings a damage probability of 0.30--0.36 against 0.77 for MambaBDA-S, so that the plug-in removes 9 of its 11 false alarms. Reliance on the surroundings thus also causes false alarms where the surroundings change for reasons other than damage, and in an unseen event building evidence corrects errors in both directions, as it does on xBD.

%% file: tables/tab_plugin.tex
\multicolumn{10}{l}{\emph{Footprints used for scoring only}} \\
ChangeOS (R101) & alone & 0.807 & 0.959 & 0.504 & 0.748 & 0.626 & 0.380 & 0.029 & \textbf{0.997} \\
 & + plug-in & \textbf{0.819} & \textbf{0.960} & \textbf{0.585} & \textbf{0.799} & \textbf{0.692} & \textbf{0.427} & \textbf{0.386} & 0.994 \\
ChangeOS (Swin-T) & alone & 0.814 & 0.959 & 0.565 & 0.754 & 0.660 & 0.405 & 0.010 & \textbf{0.999} \\
 & + plug-in & \textbf{0.823} & \textbf{0.962} & \textbf{0.594} & \textbf{0.778} & \textbf{0.686} & \textbf{0.424} & \textbf{0.130} & \textbf{0.999} \\
xView2 1st place & alone & 0.822 & \textbf{0.961} & 0.604 & 0.715 & 0.659 & 0.432 & 0.017 & \textbf{0.998} \\
 & + plug-in & \textbf{0.825} & \textbf{0.961} & \textbf{0.619} & \textbf{0.778} & \textbf{0.699} & \textbf{0.448} & \textbf{0.351} & 0.996 \\
ChangeStar (Changen2) & alone & 0.800 & 0.947 & 0.651 & 0.574 & 0.612 & \textbf{0.464} & 0.008 & 0.997 \\
 & + plug-in & \textbf{0.816} & \textbf{0.953} & \textbf{0.655} & \textbf{0.681} & \textbf{0.668} & 0.456 & \textbf{0.133} & \textbf{0.998} \\
Strong Baseline & alone & 0.795 & 0.949 & \textbf{0.679} & 0.607 & 0.643 & 0.409 & 0.054 & 0.978 \\
 & + plug-in & \textbf{0.816} & \textbf{0.958} & 0.654 & \textbf{0.721} & \textbf{0.688} & \textbf{0.436} & \textbf{0.200} & \textbf{0.996} \\
DisasterAdaptiveNet & alone & 0.795 & 0.949 & \textbf{0.653} & 0.616 & 0.635 & 0.390 & 0.021 & 0.998 \\
 & + plug-in & \textbf{0.816} & \textbf{0.956} & 0.650 & \textbf{0.718} & \textbf{0.684} & \textbf{0.443} & \textbf{0.157} & \textbf{0.999} \\
MambaBDA-S & alone & 0.813 & \textbf{0.961} & \textbf{0.614} & 0.713 & 0.663 & 0.385 & 0.060 & 0.987 \\
 & + plug-in & \textbf{0.819} & 0.960 & \textbf{0.614} & \textbf{0.774} & \textbf{0.694} & \textbf{0.433} & \textbf{0.387} & \textbf{0.995} \\
\midrule
\multicolumn{10}{l}{\emph{Footprints also used as input}} \\
Damage-TriageFormer & alone & 0.786 & 0.944 & \textbf{0.634} & 0.710 & 0.672 & 0.468 & -- & -- \\
 & + plug-in & \textbf{0.800} & \textbf{0.949} & 0.623 & \textbf{0.784} & \textbf{0.703} & \textbf{0.475} & -- & -- \\
KR-GAT & alone & 0.793 & \textbf{0.950} & 0.562 & 0.776 & 0.669 & 0.402 & -- & -- \\
 & + plug-in & \textbf{0.796} & 0.949 & \textbf{0.593} & \textbf{0.797} & \textbf{0.695} & \textbf{0.431} & -- & -- \\

%% file: tables/tab_backbone.tex
\multicolumn{7}{l}{\emph{Footprints used for scoring only}} \\
ChangeOS (R101) & 0.807 & \textbf{0.819} & 0.818 & 0.626 & 0.692 & \textbf{0.701} \\
ChangeOS (Swin-T) & 0.814 & 0.823 & \textbf{0.825} & 0.660 & 0.686 & \textbf{0.691} \\
xView2 1st place & 0.822 & \textbf{0.825} & \textbf{0.825} & 0.659 & 0.699 & \textbf{0.711} \\
ChangeStar (Changen2) & 0.800 & 0.816 & \textbf{0.818} & 0.612 & 0.668 & \textbf{0.690} \\
Strong Baseline & 0.795 & \textbf{0.816} & 0.815 & 0.643 & 0.688 & \textbf{0.697} \\
DisasterAdaptiveNet & 0.795 & \textbf{0.816} & 0.815 & 0.635 & 0.684 & \textbf{0.701} \\
MambaBDA-S & 0.813 & \textbf{0.819} & 0.818 & 0.663 & 0.694 & \textbf{0.706} \\
\midrule
\multicolumn{7}{l}{\emph{Footprints also used as input}} \\
Damage-TriageFormer & 0.786 & 0.800 & \textbf{0.801} & 0.672 & 0.703 & \textbf{0.712} \\
KR-GAT & 0.793 & \textbf{0.796} & \textbf{0.796} & 0.669 & 0.695 & \textbf{0.707} \\

%% file: tables/tab_cf_plugin.tex
ChangeOS R101 & 0.16 & 0.52 & 0.82 & 0.31 \\
ChangeOS Swin-T & 0.17 & 0.44 & 0.76 & 0.43 \\
MambaBDA-S & 0.16 & 0.52 & 0.78 & 0.32 \\
xView2 winning solution & 0.20 & 0.54 & 0.87 & 0.27 \\
ChangeStar & 0.11 & 0.42 & 0.86 & 0.44 \\
Strong Baseline & 0.21 & 0.48 & 0.78 & 0.40 \\
DisasterAdaptiveNet & 0.23 & 0.44 & 0.73 & 0.44 \\

%% file: tables/tab_mitigation.tex
Existing model & 0.785 & 0.584 & 0.754 & 0.669 \\
Retrained with xFBD augmentation & 0.773 & \textbf{0.607} & 0.726 & 0.666 \\
With the plug-in (ours) & \textbf{0.796} & \textbf{0.607} & \textbf{0.795} & \textbf{0.701} \\

%% file: tables/tab_bright.tex
UNet & alone & 0.764 & 0.973 & 0.097 & 0.895 & 0.496 & 0.082 \\
 & + plug-in & \textbf{0.774} & \textbf{0.976} & \textbf{0.100} & \textbf{0.955} & \textbf{0.527} & \textbf{0.097} \\
DeepLabV3+ & alone & 0.756 & 0.973 & \textbf{0.082} & 0.926 & 0.504 & 0.113 \\
 & + plug-in & \textbf{0.761} & \textbf{0.975} & 0.074 & \textbf{0.971} & \textbf{0.523} & \textbf{0.116} \\
SiamAttnUNet & alone & 0.786 & 0.977 & 0.101 & 0.900 & 0.501 & \textbf{0.144} \\
 & + plug-in & \textbf{0.795} & \textbf{0.981} & \textbf{0.102} & \textbf{0.956} & \textbf{0.529} & 0.133 \\
SiamCRNN & alone & 0.743 & 0.973 & \textbf{0.121} & 0.860 & 0.491 & 0.113 \\
 & + plug-in & \textbf{0.749} & \textbf{0.974} & 0.106 & \textbf{0.943} & \textbf{0.525} & \textbf{0.115} \\
DamageFormer & alone & \textbf{0.759} & 0.976 & \textbf{0.082} & 0.930 & 0.506 & 0.118 \\
 & + plug-in & \textbf{0.759} & \textbf{0.977} & \textbf{0.082} & \textbf{0.961} & \textbf{0.521} & \textbf{0.135} \\

%% file: tables/tab_turkey.tex
ChangeOS (R101) & alone & 0.346 & 0.115 & 0.014 & 0.348 & 0.111 & 0.020 & 4 \\
 & + plug-in & \textbf{0.491} & \textbf{0.192} & \textbf{0.324} & \textbf{0.507} & \textbf{0.222} & \textbf{0.354} & \textbf{1} \\
MambaBDA-S & alone & 0.481 & 0.154 & 0.319 & 0.496 & 0.167 & 0.347 & 11 \\
 & + plug-in & \textbf{0.517} & \textbf{0.218} & \textbf{0.412} & \textbf{0.535} & \textbf{0.259} & \textbf{0.463} & \textbf{2} \\
xView2 1st place & alone & 0.467 & 0.000 & 0.306 & 0.492 & 0.000 & 0.367 & 3 \\
 & + plug-in & \textbf{0.515} & \textbf{0.205} & \textbf{0.421} & \textbf{0.537} & \textbf{0.241} & \textbf{0.469} & \textbf{1} \\

%% file: sections/discussion.tex
\section{Conclusions}
\label{sec:discussion}

Building damage assessment models judge buildings largely by their neighbours. Nine models covering the main paradigms are nearly perfect on buildings that agree with their surroundings but miss a third to half of the isolated damaged buildings, and their decisions follow the surroundings when a building is moved into a new neighbourhood, including those of models that take the footprint as input. The failures stem from a neighbourhood shortcut and from the rarity of isolated damage. A building-evidence plug-in, which multiplies the probabilities of the existing model with those of a tight-crop expert and a swap-trained expert, keeps or improves the overall accuracy of all nine models and improves all of them on discordant buildings. It cuts by 40--69\% the intact buildings that a damaged neighbourhood turns into damaged ones, and it carries over to synthetic isolated damage, to optical--SAR data and to an unseen earthquake.

Because overall scores hide this behaviour, BDA benchmarks should report the accuracy on buildings that agree and disagree with their neighbourhood, and counterfactual flip rates, next to the overall score. Both need only footprints and predictions, and flood events should be reported separately.

The plug-in needs single-building footprints, from an inventory or an instance-level model, and on SAR it reduces false alarms but does not yet detect isolated damage. Future work should build building evidence into end-to-end models, apply the protocol to vision-language models used for damage assessment, and test both on conflict damage, where isolated damage is the rule rather than the exception.